\documentclass[11pt]{article}

\usepackage{acl}

\usepackage{times}
\usepackage{latexsym}

\usepackage[T1]{fontenc}

\usepackage[utf8]{inputenc}

\usepackage{microtype}

\usepackage{inconsolata}

\usepackage{graphicx}

\usepackage{amsmath} 
\usepackage{amsfonts} 
\usepackage{enumitem}
\title{KIT's Low-resource Speech Translation Systems for IWSLT2025: System Enhancement with Synthetic Data and Model Regularization}

\author{Zhaolin Li, Yining Liu, Danni Liu, Tuan Nam Nguyen, Enes Yavuz Ugan, \\ {\bf Tu Anh Dinh, Carlos Mullov, Alexander Waibel, Jan Niehues} \\ Karlsruhe Institute of Technology \\ \texttt{firstname.lastname@kit.edu}}

\begin{document}
\maketitle
\begin{abstract}
This paper presents KIT's submissions to the IWSLT 2025 low-resource track. We develop both cascaded systems, consisting of Automatic Speech Recognition (ASR) and Machine Translation (MT) models, and end-to-end (E2E) Speech Translation (ST) systems for three language pairs: Bemba, North Levantine Arabic, and Tunisian Arabic into English. Building upon pre-trained models, we fine-tune our systems with different strategies to utilize resources efficiently. This study further explores system enhancement with synthetic data and model regularization. Specifically, we investigate MT-augmented ST by generating translations from ASR data using MT models.  For North Levantine, which lacks parallel ST training data, a system trained solely on synthetic data slightly surpasses the cascaded system trained on real data. We also explore augmentation using text-to-speech models by generating synthetic speech from MT data, demonstrating the benefits of synthetic data in improving both ASR and ST performance for Bemba. Additionally, we apply intra-distillation to enhance model performance. Our experiments show that this approach consistently improves results across ASR, MT, and ST tasks, as well as across different pre-trained models. Finally, we apply Minimum Bayes Risk decoding to combine the cascaded and end-to-end systems, achieving an improvement of approximately 1.5 BLEU points.

\end{abstract}

\section{Introduction}

In this paper, we present our submissions to the IWSLT 2025 low-resource track\footnote{\url{https://github.com/ZL-KA/IWSLT25-low-resource-KIT}}. We participate in three language pairs, translating from Bemba (ISO: bem), North Levantine Arabic (ISO: apc), and Tunisian Arabic (ISO: aeb) into English. Our approach follows the unconstrained track, reflecting practical scenarios by leveraging all available resources, including multilingual pre-trained models and external datasets.

Building upon the submissions of last year \cite{li-etal-2024-kit}, which investigates efficient utilization of available resources using multilingual pre-trained models, this work explores two approaches to further enhance model performance without involving extra resources: synthetic data augmentation and model regularization.

One of the main challenges in building speech translation (ST) systems is the scarcity of end-to-end (E2E) ST data. Given that Automatic Speech Recognition (ASR) and Machine Translation (MT) resources are more accessible, we leverage them to create synthetic ST data. First, we investigate the MT-augmented approach, using a trained MT model to generate target-language translations from ASR datasets. Additionally, inspired by prior work \cite{robinson22_interspeech,10889894,eskimez2024e2ttsembarrassinglyeasy, tong2024improvinggeneralizingflowbasedgenerative, moslem-2024-leveraging}, we explore synthetic speech generation. Specifically, we train Text-To-Speech (TTS) models using ASR data and use them to generate synthesized speech from the MT datasets.

We also explore model regularization to enhance model performance. Previous research shows ST systems for low-resource languages benefit from model regularization during training because of the imbalanced parameter usage \cite{romney-robinson-etal-2024-jhu, jiawei-etal-2024-hw}. However, these works are limited to MT models in the cascaded system. Since model regularization is a generic approach, this work investigates its effectiveness with both ASR, MT, and ST tasks.

With experimental results across different language pairs, we conclude the findings as follows: 
\begin{itemize}[topsep=2pt, itemsep=1pt, parsep=1pt]
    \item Synthetic data is promising for improving model performance, provided that the generated data is of reasonable quality.
    \item Model regularization is a general approach for enhancing performance, and we demonstrate its effectiveness across different tasks and pre-trained models.
    \item The various differences between languages and corpora lead to divergent findings in terms of pre-trained model effectiveness and training strategies, highlighting the need for language-specific approaches.
\end{itemize}

\section{Task Description} \label{s2}

The IWSLT 2025 low-resource track defines two system categories: constrained, where models are trained exclusively on datasets provided by the organizers, and unconstrained, where participants are free to use any external resources. In this work, we focus on the unconstrained condition, aiming to reflect better practical and real-world scenarios, where leveraging diverse data sources is often essential for building effective translation systems.

\subsection{Development Dataset}

This work focuses on three language pairs with the source languages of Bemba, North Levantine, or Tunisian, and the same target language of English. The development data used for these tasks is summarized in Table \ref{tab:dev_data}. Notably, North Levantine lacks end-to-end parallel training data, highlighting the need for additional resources and data augmentation techniques to build effective translation models for this language.

\begin{table}[h]
    \centering
    \begin{tabular}{cccc} \hline
         & Train & Valid & Test \\ \hline
    apc  & - & 1126 & 975  \\
    aeb & 202k & 3833 & 4204 \\
    bem & 82k & 2782 & 2779 \\ \hline
    \end{tabular}
    \caption{Statistics on development data. The value indicates the number of samples, where one sample is composed of the audio, transcript in the language, and translation in English.}
    \label{tab:dev_data}
\end{table}

\subsection{Additional Dataset}

Under the unconstrained condition, we utilize additional resources to improve model performance, as detailed in Table \ref{tab:extra_data}. These supplementary datasets include ASR and MT datasets, but notably no end-to-end ST dataset due to unavailability. This highlights the advantages of building cascaded ST systems, which can effectively leverage separate ASR and MT components. All additional datasets are publicly accessible, except SyKIT and MINI, which are internally developed and originate from conversational speech data.

\begin{table}[h]
\centering
\begin{tabular}{ccccc}
\hline
Lang. & Corpus & Type & Amount. \\ \hline

apc& LDC2005S08  & ASR & 60h  \\ 
& LDC2006S29 & ASR & 250h \\ 
& SyKIT & ASR & 50h \\
& Tatoeba & MT & 20  \\ 
& UFAL & MT & 120k \\ 
& LDC2012T09 & MT & 138k \\ \hline

aeb& SRL46 & ASR & 12h \\ 
& GNOME & MT & 646  \\ \hline

ara& SLR148 & ASR & 111h  \\
& MGB & ASR & 1200h \\
& MINI & ASR & 10h  \\
& CCMatrix & MT & 5M \\
& NLLB & MT & 5M  \\
& OpenSubtitles & MT & 3M \\ \hline

bem& BembaSpech & ASR &  24h \\

& NLLB & MT & 427k \\ \hline

\end{tabular}
\caption{Overview of the additional data resources. The unit in amount is the number of hours or sentences.}
\label{tab:extra_data}
\end{table}

\section{Approaches}

\subsection{Synthetic Data Augmentation}

Data scarcity remains a key challenge in low-resource natural language processing tasks, particularly for end-to-end speech translation (ST). To address this limitation, this work investigates data augmentation approaches using synthetic data. We focus on two augmentation approaches that address different modalities: the MT-augmented method, which generates synthetic translations from ASR data, and the TTS-augmented method, which produces synthetic speech from MT data. Together, these methods aim to enhance the quality and robustness of ST models in low-resource settings.

\subsection{Model Regularization}

Regularization remains a simple yet powerful way to boost the generalisation capacity of neural sequence models, and has already proved valuable in machine translation through techniques such as R‑Drop and its variants \cite{wu2021r,xu2022importance}. Motivated by the recent success of intra‑distillation (ID) in low-resource MT \cite{romney-robinson-etal-2024-jhu}, we extend ID to all three tasks: ASR, MT, and ST, based on the public implementation with the following modification\footnote{\url{https://github.com/fe1ixxu/Intra-Distillation/}}.

Unlike previous work that directly fine-tunes a pre‑trained model with a loss that combines the task objective and ID, we notice that direct fine-tuning leads to suboptimal performance in preliminary experiments. We therefore adopt a two‑stage approach: (1) vanilla fine‑tuning to adapt the pre‑trained model to the downstream task, followed by (2) ID fine‑tuning to regularize the adapted model with its own intermediate predictions. This simple approach retains the advantages of task‑specific adaptation while unlocking the additional robustness that ID provides.

\subsection{System combination}

Following the prior work \cite{li-etal-2024-kit}, we combine the cascaded system and the end-to-end system with Minimum Bayes Risk (MBR) decoding to boost model performance \cite{kumar2004minimum}. Specifically, with 50 hypothesis from the cascaded system and 50 from the end-to-end system as the pseudo-references,  we use the official evaluation metric BLEU as the utility function in our MBR decoding.

\subsection{Arabic Dialects Normalization}
This work focuses on ST tasks, where normalizing intermediate transcripts can streamline the overall process. Following the approach proposed by \cite{ben-kheder-etal-2024-aladan}, we implement a dialect-specific normalization pipeline to ensure consistent pre-processing across diverse transcriptions in North Levantine and Tunisian dialects. Our normalization process includes compound word splitting, orthographic normalization of dialectal variations, and numeral normalization.

\section{Experimental Setups and Results}

\subsection{Preprocessing}

Following prior work \cite{li-etal-2024-kit}, we exclude speech segments exceeding 15 seconds in duration due to computational limitations. Subsequently, we apply speech augmentation techniques including Gaussian noise injection, time stretching, time masking, and frequency masking.

\subsection{Pre-trained Models}

In this work, we explore fine-tuning with the following pre-trained models for different tasks.

\textbf{SeamlessM4T}: SeamlessM4T \cite{barrault2023seamlessm4t} is a highly multilingual and multimodal model that has demonstrated strong performance in low-resource scenarios across ASR, MT, and ST tasks. We use the large configuration of version 2 for our experiments\footnote{\url{https://huggingface.co/facebook/seamless-m4t-v2-large}}. It is important to note that none of the three source languages used in our experiments were included in SeamlessM4T's pre-training data.

\textbf{NLLB}: NLLB \cite{costa2022no} is a multilingual machine translation model capable of directly translating between 200 languages. Its pre-training data includes a wide range of languages, particularly many low-resource ones, making it well-suited for low-resource translation tasks. North Levantine and Tunisian are included in its pre-training, and Bemba is not. 

We use the 1.3B parameter version\footnote{\url{https://github.com/facebookresearch/fairseq/tree/nllb}}, freezing the word embeddings to reduce memory usage. We also freeze the decoder except for the cross-attention layers, as suggested in \cite{cooper-stickland-etal-2021-recipes}. Due to the lack of MT data for North Levantine, we fine-tune the model jointly on Tunisian and Modern Standard Arabic, resulting in many-to-English MT systems.

\textbf{MMS}: MMS is a multilingual speech recognition model pre-trained on data from over 1,100 languages. Its broad language coverage and use of self-supervised learning enable effective fine-tuning for low-resource languages. For our experiments, we add a linear layer on top of the pre-trained encoder and fine-tune the model using the CTC loss\footnote{\url{https://huggingface.co/facebook/mms-300m}}. Additionally, we explore enhancements through shallow fusion with language models using different tokenization strategies \cite{li-niehues-2025-enhance}.

\textbf{XEUS}: Similar like MMS, XEUS is a multilingual encoder-based speech recognition model\cite{chen2024towards}. It is pre-trained on approximately 1 million hours of unlabeled audio spanning 4,057 languages. Moreover, it incorporates dereverberation training, enhancing its robustness to various acoustic conditions. We apply the same fine-tuning strategy used for MMS to XEUS\footnote{\url{https://huggingface.co/espnet/xeus}}.

\subsection{Synthetic Data}

We explore two TTS systems, each is optimized for different strengths.

\subsubsection{E2TTS}

E2TTS \cite{eskimez2024e2ttsembarrassinglyeasy} is a recent non-autoregressive text-to-speech (TTS) model that demonstrates strong performance. Unlike previous non-autoregressive approaches, it upsamples the text sequence to the spectrogram length by padding, which eliminates the need for explicit monotonic alignment search and duration modeling during training. This simplifies the training process and makes the model more end-to-end. Besides, E2TTS utilizes conditional flow matching \cite{tong2024improvinggeneralizingflowbasedgenerative} as its backbone, inheriting its strong generative capabilities that ensure the naturalness and high-fidelities of the synthesized audio.

Additionally, its combination of in-context learning and classifier-free guidance \cite{ho2022classifierfreediffusionguidance} enables highly flexible zero-shot synthesis. This means we can generate audio using a randomly given audio prompt that indicates the target speaker’s identity, emotion tone, background noise profile, etc, and we could also control how much of these acoustic characteristics from the prompt would be bypassed to model output. These features allow us to create more diverse audio samples ideal for data augmentation.

As for training configurations, we use a checkpoint pretrained 
on English as a startup. We follow training hyperparameters from the original paper with modified vocabulary size tailored to our target languages and datasets. Additionally, we use Vocos \cite{siuzdak2024vocosclosinggaptimedomain} vocoder to synthesize waveforms from log mel-filterbank features.

Following model training, we synthesize audio samples for data augmentation by running inference on source transcripts. For each generation, we condition the model using a randomly selected text-audio pair from the training dataset as a prompt, employing classifier-free guidance \(\alpha=2.0\) to strengthen prompt adherence. This ensures that the speaker distribution in the generated data matches that of the original dataset. Additionally, we configure the numerical approximation steps to 32 to ensure high-quality waveform generation.

\subsubsection{VITS}

VITS~\cite{kim2021conditional} is a conditional variational autoencoder architecture enhanced with normalizing flows. It comprises three primary components: a posterior encoder, a prior encoder, and a waveform generator. These modules respectively model the distributions \( q_\phi(z|x) \), \( p_\theta(z|c) \), and \( p_\psi(y|z) \). Specifically, \( q_\phi(z|x) \) represents the posterior distribution, and \( p_\psi(y|z) \) corresponds to the data distribution, with parameters learned by the posterior encoder \( \phi \) and the HiFi-GAN waveform generator \( \psi \)~\cite{NEURIPS2020_c5d73680}. Here, \( x \) denotes the speech input, \( z \) is the latent variable, and \( y \) is the resulting waveform. The prior distribution \( p_\theta(z|c) \), parameterized by the prior encoder \( \theta \), is further refined using a normalizing flow \( f \), where the latent variables are conditioned on the text input \( c \).

During training, the model is optimized to maximize the conditional likelihood \( p(x|c) \) by maximizing its evidence lower bound (ELBO):

\vspace{-3mm}
\begin{equation}
\begin{split}
\log p(x|c) \geq 
& \ \mathbb{E}_{q_\phi(z|x)}[\log p_\psi(x|z)] \\
& \ - D_{\text{KL}}(q_\phi(z|x) \| p_\theta(z|c))
\label{eq3}
\end{split}
\end{equation}

We train the model from scratch and fine-tune it for 1,000,000 steps using a setup similar to that in the original VITS paper. After training, we synthesize audio samples for data augmentation by performing inference on the source transcripts. For each synthesized audio, a random speaker is selected from the training set, which includes approximately 75 speakers, to produce diverse speaker-conditioned outputs.

\subsection{Evaluation Metrics}

Following the evaluation instruction of IWSLT 2025 low-resource track, both prediction and reference are lowercased and punctuation removed\footnote{\url{https://github.com/kevinduh/iwslt22-dialect}}. We use Character Error Rate (CER) and Word Error Rate (WER) as ASR evaluation metrics. For translation tasks, we use evaluation metrics of Bilingual Evaluation Understudy (BLEU) and Character n-gram F-score (chrF).

\begin{table}[h]
\centering
\begin{tabular}{cccc} \hline

ID & Model &  bem\_valid & bem\_test  \\ \hline
A1 & MMS & 10.8/40.4 & 10.0/37.3   \\
A2 & A1 + LM & 9.8/36.6 & 8.8/34.8 \\
A3 & XEUS & 10.7/41.0 & 10.0/39.4 \\

A4 & Seamless &  10.8/37.1 & 10.0/36.6 \\
A5 & Seamless all & 10.0/34.1 & 9.3/33.1 \\
A6 & A5 + ID & \textbf{9.8/33.1} & \textbf{9.1/31.9} \\

\hline

B1 & NLLB all & 26.0/51.0 & 28.6/52.4  \\
B2 & NLLB & 25.6/51.5 & 28.5/52.6 \\
B3 & B1 + ID & 27.1/52.0 & 29.1/52.6 \\

B4 & Seamless all & 26.6/52.8& 26.8/52.3 \\
B5 & Seamless & 27.9/52.3 & 27.9/52.6\\
B6 & B5 + ID & \textbf{28.6/54.7} & \textbf{29.3/54.5} \\

\hline

C & Best A+B & 28.4/53.0 & 28.9/52.8\\

\hline

D1 & Seamless & 27.6/51.1 & 27.7/51.3\\
D2 & D1 + ID & \textbf{29.5/53.6} & \textbf{29.8/53.1} \\
D3 & D1 + TTS & 28.0/52.6 & 28.7/53.0\\
D4 & D3 + ID & 29.4/53.6 & 29.3/53.3\\

\hline

E1 & C & 29.4/52.0& 29.0/51.5\\
E2 & D4 & 30.0/52.7& 29.8/52.3 \\
E3 & E1 + E2 & \textbf{31.1/53.4} & \textbf{30.8/52.9}\\
\hline

\multicolumn{2}{l}{Best ST system 2024} & 26.3/- & 30.4/- \\
\hline
\end{tabular}
\caption{Experimental results for Bemba to English. \textbf{A} indicates ASR systems, \textbf{B} indicates MT systems with gold transcript, \textbf{C} indicates cascaded systems, \textbf{D} indicates  E2E ST systems, and \textbf{E} indicate MBR systems. \textbf{all} indicates training with all available resources; otherwise, training is done with only the development resource. ASR results are reported as CER/WER, while MT and ST results are presented as BLEU/chrF.}
\label{tab:bem_results}
\end{table}

\begin{table*}[h]
\centering
\begin{tabular}{ccccccc} \hline

ID & Model & apc\_valid & apc\_test & aeb\_valid & aeb\_test  \\ \hline

A1 & Seamless all ara & 45.1/68.4 & 12.4/37.5 & 18.2/36.8 & 23.2/44.5 \\
A2 & A1 + transfer & \textbf{45.0/66.7} & \textbf{12.0/37.0} & \textbf{18.4/36.9} & \textbf{21.7/41.3} \\

A3 & A2 + ID & 47.9/70.1 & 16.1/42.8 & 19.6/39.4 & 22.7/43.5 \\

\hline

B1 & NLLB all & 24.9/53.6 & 20.9/48.8 & 30.4/52.6 & 26.8/50.2\\
B2 & B1 + transfer & \textbf{31.3/57.6} & \textbf{28.0/54.4} & \textbf{30.3/52.2} & \textbf{26.3/49.9}\\ 
B3 & Seamless & 21.7/48.2 & 18.9/45.1 & 28.4/50.8 & 25.6/48.9\\

\hline

C & Best A+B &   19.1/42.1 &26.6/53.2 & 23.4/46.2 & 20.1/43.8\\

\hline

D1 & Seamless & 19.9/41.7&27.3/52.4 & 20.5/43.3 & 18.0/41.1\\
D2 & Seamless + ID  & - & - & \textbf{22.9/45.4} & \textbf{19.6/43.8} \\

\hline
E1 & C & 19.0/41.4 & 26.5/52.6 & 23.4/46.2 & 20.2/43.4 \\
E2 & Best D & 19.7/41.1 & 27.4/51.9 & 23.1/45.2 & 19.9/42.5 \\
E3 & E1+E2 & \textbf{21.0/42.5} & \textbf{29.4/53.8} & \textbf{24.6/46.9} & \textbf{21.3/44.4} \\
\hline
\multicolumn{2}{l}{Best ST system 2024/2023} & 26.9/51.9 & 28.7/52.3 & 24.9/- & 22.2/- \\
\hline

\end{tabular}
\caption{Experimental results for North Levantine and Tunisian to English. \textbf{A} indicates ASR systems, \textbf{B} indicates MT systems with gold transcript, \textbf{C} indicates cascaded systems, \textbf{D} indicates  E2E ST systems, and \textbf{E} indicate MBR systems. \textbf{all} indicates training with all available resources; otherwise, training is done with only the development resource. \textbf{transfer} indicates a second-step fine-tuning. ASR results are reported as CER/WER, while MT and ST results are presented as BLEU/chrF.}
\label{tab:ara_results}
\end{table*}

\subsection{ASR Systems} \label{sec:s4_asr}

Due to limitations in time and computational resources, we primarily experiment with ASR systems for Bemba. The corresponding results, identified by IDs starting with 'A' in Table \ref{tab:bem_results}, are discussed below. In experiments A1 and A2 using MMS, we observe that applying language model fusion with encoder-based models consistently improves ASR performance, resulting in a reduction of approximately 4 WER points—aligning with findings from prior work. Comparing A1 and A3, we observe that XEUS achieves performance similar to MMS, despite being pre-trained on more languages and incorporating dereverberation augmentation. The possible explanations are that the audios are recorded in controlled conditions with minimal background noise, and the additional language coverage of XUES pre-training benefits little to Bemba in terms of speech representation.

Compared to the encoder-only models above, the encoder-decoder model SeamlessM4T achieves comparable performance when fine-tuned using only development resources. We apply several training strategies to SeamlessM4T: specifically, we compare using only the development resources versus all available resources with the pre-trained model. As seen from A4 to A5, utilizing all resources results in about a 3-point WER improvement. Furthermore, we achieve an additional improvement of approximately 1 WER point by applying ID on top of A5.

For Arabic dialects, we first fine-tune with all resources, including MSA,  with SeamlessM4T, then fine-tune with the datasets of the target language pairs in the second stage. This benefits in tackling the limited training resources under the normalization processing, which brings the dialects and standard similar in terms of learning speech representation. As Table \ref{tab:ara_results} shows, the transfer learning slightly improves model performance. Notably, the ASR systems for North Levantine have unbalanced results for validation and test splits, despite the test split remaining untouched during training. One hypothesis is a domain mismatch between these splits. Further investigation is needed to confirm this hypothesis.

\subsection{MT Systems}

We experimented with SeamlessM4T and NLLB models, chosen for their differing language coverage and capabilities. Two fine-tuning strategies were explored: one using all available resources followed by transfer to the development set, and another using only the development set for fine-tuning.

For Bemba, fine-tuning exclusively on the development dataset yielded better performance than using all resources, as shown in Table \ref{tab:bem_results}. The choice of fine-tuning resources had little effect on NLLB's performance. When comparing pre-trained models, NLLB outperformed SeamlessM4T, under the condition that Bemba is included in the pretraining data of either model. Notably, incorporating ID data improved MT performance for both models by approximately 1 BLEU point.

For North Levantine and Tunisian, we experiment with NLLB fine-tuning using all Arabic resources, followed by a second-step fine-tuning with only the available resources for each language pair, for the same reasons as in Section \ref{sec:s4_asr}. Specifically, we fine-tune with the UFAL and LDC2012T09 datasets for North Levantine and the development dataset for Tunisian in the second-step fine-tuning, based on availability. We observe a significant improvement for North Levantine, consistent with \cite{ben-kheder-etal-2024-aladan}, potentially due to the benefits of domain similarity. In contrast, the performance with second-step fine-tuning slightly declines for Tunisian. This underscores the importance of language-specific approaches. 

We also fine-tune the pre-trained SeamlessM4T using only the development set and find that its performance falls noticeably behind that of NLLB, though the comparison is not entirely fair. Given NLLB’s pre-training advantage on these languages and the preliminary results, we did not apply the same fine-tuning strategy for SeamlessM4T due to time limitations.

\subsection{Synthetic Data Augmentation}

As described in Section \ref{s2}, there is no E2E ST training data available for North Levantine. To address this, we explore synthetic data augmentation using both MT-augmented and TTS-augmented approaches to create ST training data. In addition, we also apply the TTS-augmented approach to Bemba to examine the impact of additional synthetic ST data.

\subsubsection{MT-augmented ST systems}

Using the MT system B2 in Table \ref{tab:ara_results}, we generated translations from the ASR dataset LDC2005S08 (listed in Table \ref{tab:extra_data}) to create synthetic ST data. After applying filtering criteria such as the audio-to-text length ratio, the generation ends with 45K samples. We then train E2E ST systems with the SeamlessM4T model using only the synthetic data for training and the validation split of the development set for validation. As shown in Table \ref{tab:mt_sys_results}, the performance of the ST systems relates to the volume of data used, highlighting the importance of selecting an appropriate amount of synthetic data. 

Notably, the best-performing ST system trained on synthetic data surpasses the cascaded system, which is trained with real ASR and MT data, by approximately 1 BLEU point. This improvement may be attributed to the robustness of the MT system, which generates reasonably accurate synthetic translations.


\begin{table}[]
    \centering 
    \begin{tabular}{ccc} 
    \hline
   \#Synthetic data & Valid & Test \\ \hline
    45K & 19.1/41.3 & 26.2/51.9  \\
    23K & \textbf{19.9/41.7} & \textbf{27.3/52.4} \\
    12K & 19.7/41.4 & 27.3/52.4 \\
    6K & 19.2/41.4 & 26.6/52.4 \\

\hline

Cascaded & 19.1/42.1 & 26.6/53.2 \\ \hline
    
    \end{tabular}
    \caption{MT-augmented ST systems for North Levantine. The results are presented as BLEU/chrF.}
    \label{tab:mt_sys_results}
\end{table}

\subsubsection{TTS-augmented ST systems}

For Bemba, we explore the use of ViTTS and E2TTS to generate synthetic training data. The TTS models are trained using the training split of the development dataset. The source text used for synthesis is derived from NLLB, selected based on criteria such as appropriate text length, as outlined in Table \ref{tab:extra_data}. Evaluation results for the TTS systems are provided in Appendix \ref{app_tts}.

We generate 120K synthetic training samples for each TTS model. This synthetic data is combined with the original development set for training, while the validation split remains unchanged. Following the procedure used for other end-to-end speech translation systems, we fine-tune the pre-trained SeamlessM4T models. As shown in Table \ref{tab:tts_bigc_result}, the inclusion of synthetic samples yields an improvement of up to one BLEU point compared to training without them. The quantity of synthetic data appears to affect performance; however, no consistent trend is observed regarding the optimal amount.

\begin{table}[h]
    \centering
    \begin{tabular}{cccc} \hline
    & 30K & 60K & 120K \\ \hline 
    VITS & 28.0/52.6 & 28.6/52.7 & 28.3/52.6 \\
    E2TTS & 28.7/53.0 & 28.5/52.8 & 28.3/52.7 \\ \hline
    No TTS &  \multicolumn{3}{c}{27.7/51.3}\\
    \hline
    \end{tabular}
    \caption{TTS-augmented ST systems for Bemba with scores on the test split. The column name indicates the number of synthetic data. The results are presented as BLEU/chrF.}
    \label{tab:tts_bigc_result}
\end{table}

We also explored generating synthetic ST data for North Levantine, for which no end-to-end ST data is available. We select the E2TTS model for this setting, based on its marginally better performance observed in the Bemba experiments. The training data for the TTS model comes from the ASR dataset LDC2005S08, while the MT dataset UFAL is used for speech generation. This process yields 60K ST samples, selected using the same criteria as in the Bemba experiments. Given the lack of end-to-end ST training data for North Levantine, we examine training solely with synthetic data, using real data only for validation. As shown in Table \ref{tab:tts_apc_result}, relying exclusively on synthetic data results in lower performance compared to the cascaded system. We attribute this to the under-developed TTS model, as reflected in its evaluation in Appendix \ref{app_tts}.

\begin{table}[h]
    \centering
    \begin{tabular}{ccc} \hline
    \#Synthetic data & Valid & Test \\ \hline 
    
    60K & 9.6/29.2 & 12.9/35.5 \\
    30K & 9.2/28.6 & 11.9/34.5 \\
    15K & 10.8/30.6 & 13.5/36.8 \\ \hline

    Cascaded & 19.1/42.1 & 26.6/53.2 \\
    \hline
    \end{tabular}
    \caption{TTS-augmented ST systems for North Levantine. The results are presented as BLEU/chrF.}
    \label{tab:tts_apc_result}
\end{table}

\subsection{Regularization Enhancement}

We conduct experiments with ID across various systems, spanning different tasks and pre-trained models, and consistently observe performance gains. Specifically, ID leads to approximately a 1-point WER reduction in ASR and around a 1 BLEU point gain in both MT and ST tasks. However, we note an exception: ID negatively impacts ASR performance for Arabic dialects. Further investigation is needed to understand the underlying causes of this issue.

Additionally, we find that regularization enhancement and synthetic data augmentation can be additive. Adapting a model trained on synthetic data with ID yields further improvements, as illustrated by the D4 row in Table \ref{tab:bem_results}.

\subsection{Cascaded VS E2E Systems}

We compare the performance of these two widely used and distinct ST systems in low-resource scenarios, but the results are mixed and show no consistent trend. For Bemba and North Levantine, end-to-end systems outperform cascaded systems by approximately 1 BLEU point. In contrast, for Tunisian, end-to-end systems slightly underperform, with a gap of around 0.5 BLEU points. These varying results underscore the importance of adopting language- and dataset-specific strategies in low-resource speech translation.

\subsection{MBR Decoding}

We apply MBR decoding to the cascaded systems, the E2E systems, and their combination. As presented in Tables \ref{tab:bem_results} and \ref{tab:ara_results}, MBR decoding consistently yields minimal to no improvement when applied to individual systems. In contrast, combining the cascaded and E2E systems with MBR decoding consistently results in an improvement of approximately 1.5 BLEU points.

\subsection{Submission}

The same submission strategy is applied across all three language pairs. The primary system is the MBR combination of the cascaded and E2E systems. The E2E and cascaded systems are the contrastive 1 and 2 systems, respectively.

Table \ref{tab:official_eval} presents the evaluation results reported by \citet{abdulmumin-etal-2025-findings}. The test data includes two datasets (test2022 and test2023) for Tunisian and one dataset each for North Levantine and Bemba. Referring to the previous results, the performance comparison between cascaded and E2E systems remains consistent for Bemba, with the cascaded system outperforming the E2E system. In contrast, opposite trends are observed for the Arabic dialects. This difference underscores the necessity for language- or corpus-specific analyses. The MBR combination of cascaded and E2E systems consistently yields performance improvements, highlighting the advantage of integrating both systems.

\begin{table*}[]
    \centering
    \begin{tabular}{ccccc} \hline
         & aeb test22 & aeb test23 & apc & bem \\
    ASR     & 21.0/40.5   & 23.0/41.8 & - & 9.2/31.9\\
    ST Primary& 22.7/44.4 & 21.4/42.3 & 23.3/45.1 & 30.3/-\\
    ST contrastive1& 21.2/43 & 19.3/40.9 & 19.1/41.0 & 29.7/-\\
    ST contrastive2 & 21.4/43.7 & 19.2/41.1 & 21.9/44.7 & 28.8/-\\
    \hline
    \end{tabular}
    \caption{Evaluation results of the submission. The ASR systems are evaluated with CER/WER. The ST systems are evaluated with BLEU/chrF.}
    \label{tab:official_eval}
\end{table*}

\section{Conclusion}

We participate in the IWSLT 2025 low-resource track, focusing on three language pairs with Bemba, North Levantine, and Tunisian as source languages, and English as the target language. Our focus is on improving model performance through synthetic data augmentation and model regularization. The results demonstrate that high-quality synthetic data can significantly enhance performance. In addition, model regularization proves to be a robust and broadly effective approach across all ASR, MT, and ST tasks in low-resource settings. Finally,  our findings highlight the importance of language-specific strategies for building effective speech translation systems, as reflected in the varying outcomes observed across the three language pairs.

\paragraph{Acknowledgement}

This work is partly supported by the Helmholtz Programme-oriented Funding, with project number 46.24.01, named AI for Language Technologies, funding from the pilot program Core-Informatics of the Helmholtz Association (HGF). It is also supported by the Deutsche Forschungsgemeinschaft (DFG) under the project Computational Language Documentation by 2025 (CLD 2025). Additional support is from the Federal Ministry of Education and Research (BMBF) of Germany under the number 01EF1803B (RELATER). 

The work was partly performed on the HoreKa supercomputer funded by the Ministry of Science, Research and the Arts Baden-Württemberg and by the Federal Ministry of Education and Research. 

\bibliography{custom}

\appendix

\section{TTS evaluation} \label{app_tts}

To evaluate the articulation quality of the trained TTS models, we used two metrics: MCD\footnote{\url{https://github.com/ttslr/python-MCD?tab=readme-ov-file}} (Mel-Cepstral Distortion \cite{Kubichek1993MelcepstralDM}) and WER. We compute MCD by first extracting 26-dimensional mel-cepstral coefficients from both synthesized and ground-truth speech samples in the validation dataset. To address temporal mismatches between sequences, we employ dynamic time warping (DTW) \cite{salvador2007fastdtw} to align the synthesized and reference feature trajectories. The final MCD metric is calculated using the 1-25th coefficients (excluding the energy term) across DTW-aligned frames.


Additionally, since MCD is not a speaker-independent metric like WER, to reduce the influence of speaker attributes, we conducted assessments in both same-speaker (reconstruction) and cross-speaker settings. The results in Table \ref{tab:tts_eval} show that trained TTS models are able to accurately reconstruct the ground-truth audio. In the cross-speaker setting, the MCD scores increase as expected but remain within a reasonable range.

For WER evaluation we use two ASR models trained without the augmented TTS data. Specifically, we use model A5 from Table \ref{tab:bem_results} for Bemba and model A2 from Table \ref{tab:ara_results} for North Levantine. As presented in Table \ref{tab:tts_eval}, E2TTS achieves reasonable WER performance for low-resource language Bemba, especially considering that the ASR system reports a WER of 31.9 on real data. In contrast, the VITS model underperforms relative to E2TTS in WER evaluations, consistent with the results in Table \ref{tab:tts_bigc_result}. 

As for low-resource language North Levantine, the WER scores are considerably high, suggesting that the E2TTS model remains underdeveloped. This likely contributes to the poor performance of ST models trained with TTS-augmented data, as indicated in Table \ref{tab:tts_apc_result}. Further analysis is needed to better understand this underdeveloped TTS model.

\begin{table}[t]
    \centering
    \begin{tabular}{llcc} \hline
     &  MCD & WER \\ \hline
    Bemba \\ \hline
  VITS same speaker & 5.4 & 51.0\\
  E2TTS same speaker  & 5.6 & 40.9\\
  E2TTS cross speaker & 7.7 & 41.9 \\ \hline

  North Levantine \\ \hline
  
  E2TTS same speaker & 4.2 & 113.3 \\
  
  E2TTS cross speaker  & 9.0 & 108.3 \\ \hline
  
    \end{tabular}
    \caption{TTS system evaluation.}
    \label{tab:tts_eval}
\end{table}

\end{document}